\documentclass[10pt,twocolumn,letterpaper]{article}

\usepackage{cvpr}              

\usepackage[dvipsnames]{xcolor}

\definecolor{cvprblue}{rgb}{0.21,0.49,0.74}
\usepackage[pagebackref,breaklinks,colorlinks,citecolor=cvprblue]{hyperref}

\usepackage{multirow}
\usepackage{float}
\usepackage{pifont}
\usepackage{svg}
\usepackage{algorithm}
\usepackage{algpseudocode}

\usepackage{graphicx}
\usepackage{amsmath}
\usepackage{amssymb}
\usepackage{tikz}
\usepackage{textcomp}
\usepackage{xcolor}
\usepackage{booktabs}
\usepackage[accsupp]{axessibility}
\def\paperID{41} 
\def\confName{CVSPORTS}
\def\confYear{2024}

\begin{document}

\title{PitcherNet: Powering the Moneyball Evolution in Baseball Video Analytics}

\author{Jerrin Bright\hspace{0.75em} 
Bavesh Balaji\hspace{0.75em} 
Yuhao Chen\hspace{0.75em} 
David A Clausi\hspace{0.75em} 
John S Zelek\hspace{0.75em} \\
{\tt\small \{jerrin.bright, bbalaji, yuhao.chen1, dclausi, jzelek\}@uwaterloo.ca}\\
University of Waterloo, Waterloo, ON, Canada, N2L 3G1\\
}

\twocolumn[{%
\renewcommand\twocolumn[1][]{#1}%
\maketitle
\begin{center}
    \vspace{-0.25in}
    \centerline{
   \includegraphics[width=\textwidth]{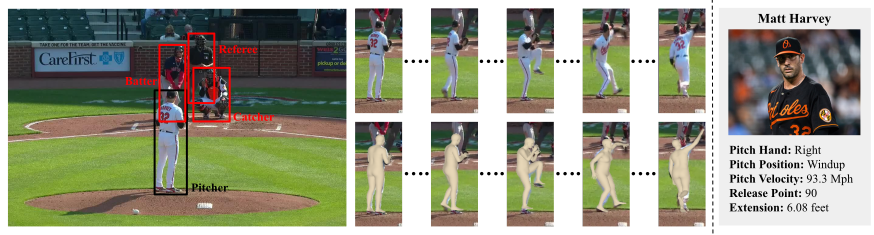}
     }
    \vspace{-0.165in}
   \captionof{figure}{\textbf{3D player reconstruction and kinematic-driven pitch statistics from monocular video.} We introduce \textit{PitcherNet}, a pioneering deep learning system that tackles low-resolution video limitations through efficient 3D human modeling for robust player alignment (left) and reliable pitch statistics analysis from estimated kinematic data (right).
   }
   \vspace{-0.05in}
\label{fig:teaser}
\end{center}%
}]

\begin{abstract}
    In the high-stakes world of baseball, every nuance of a pitcher's mechanics holds the key to maximizing performance and minimizing runs. Traditional analysis methods often rely on pre-recorded offline numerical data, hindering their application in the dynamic environment of live games. Broadcast video analysis, while seemingly ideal, faces significant challenges due to factors like motion blur and low resolution. To address these challenges, we introduce PitcherNet, an end-to-end automated system that analyzes pitcher kinematics directly from live broadcast video, thereby extracting valuable pitch statistics including velocity, release point, pitch position, and release extension. This system leverages three key components: (1) Player tracking and identification by decoupling actions from player kinematics; (2) Distribution and depth-aware 3D human modeling; and (3) Kinematic-driven pitch statistics. Experimental validation demonstrates that PitcherNet achieves robust analysis results with 96.82\% accuracy in pitcher tracklet identification, reduced joint position error by 1.8mm and superior analytics compared to baseline methods. By enabling performance-critical kinematic analysis from broadcast video, PitcherNet paves the way for the future of baseball analytics by optimizing pitching strategies, preventing injuries, and unlocking a deeper understanding of pitcher mechanics, forever transforming the game. 
\end{abstract}    
\section{Introduction}\label{sec:intro}

Driven by sabermetrics, pioneered by the Society of American Baseball Research (SABR) \cite{sabermetric_research}, baseball analytics has transformed the sport into a data-driven powerhouse, revolutionizing player evaluation, strategic decision-making, and the pursuit of victory \cite{mizels2022current}. One crucial aspect of this transformation is the analysis of pitch mechanics, where subtle movements such as strides, arm angles, and ball release points significantly affect performance \cite{kelly2019sabermetrics, karnuta2020machine}. Analyzing these intricate actions goes beyond traditional pitch type classification (fastball, curveball, etc.), delving into metrics that contribute to strategic deception, such as windup styles, varying velocities, induced ball movement, and ball release point.

Current research on baseball game analysis often rely on numerical databases containing pre-recorded offline data \cite{bock2015pitch, hickey2020dissecting, sidle2018pred, woodward2014decision}. These methods typically focus on predicting actions or game statistics based on these historical records. While some approaches utilize real-time data, they are often limited to controlled laboratory environments with expensive motion-capture setups \cite{manzi2024pitch, oyama2017, scarborough2021association}. This restricts the generalizability of their findings to the dynamic and complex situations encountered during live games. Live game broadcasts, however, offer a more holistic perspective by capturing the entirety of a pitcher's motion within the game's natural environment. This approach overcomes the limitations of controlled settings. However, analyzing broadcast data presents its own challenges, such as motion blur and low video resolution, which can significantly hinder accurate pitch analysis and potentially lead to unreliable results.

To bridge the limitations of existing methods and address the challenges of analyzing real-world live broadcasts, we introduce PitcherNet, an end-to-end automated system designed to predict performance-critical pitch statistics from the kinematic data derived from live broadcast videos. PitcherNet transcends existing approaches by meticulously analyzing each stage of the pitcher's movement, from player identification to pose estimation, and finally pitch analysis. Some crucial pitch statistics that PitcherNet estimates include pitch position, pitch velocity, ball release point, and release extension. Human mesh recovery and pitch statistics derived from the PitcherNet system are illustrated in Figure \ref{fig:teaser}. To the best of our knowledge, this is the only system, that extracts pitch statistics extensively driven from the pitcher kinematics from low-quality broadcast videos. The primary contributions of this paper includes the following:

\begin{enumerate}
    \item We introduce PitcherNet, a novel automated system, which enables accurate prediction of baseball pitch statistics from low-quality broadcast videos.
    \item We propose an innovative pitcher identification strategy which aims in classifying players by decoupling actions from player kinematics. 
    \item Building on D2A-HMR, we improve upon the modeling technique by incorporating transformers with motion blur augmentation and additional regularization heads.
    \item We demonstrate state-of-the-art results on a low-quality baseball dataset, MLBPitchDB, validating its applicability in challenging visual environments.
\end{enumerate}

\begin{figure*}[t]
  \centering
  \begin{tikzpicture}
    \node at (0,0) {\includegraphics[width=\linewidth]{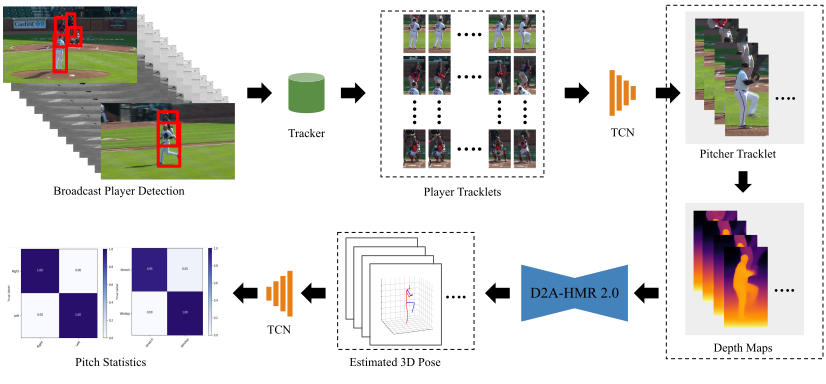}};
    \node[font=\tiny] at (-0.42,3.25) {$\mathcal{T}_1$};
    \node[font=\tiny] at (-0.43,2.5) {$\mathcal{T}_2$};
    \node[font=\tiny] at (-0.41,0.9) {$\mathcal{T}_3$};
    \node[font=\tiny] at (8.12,0.8) {$(\mathcal{T}_p)$};   
    \node[font=\tiny] at (1.00,-3.59) {$(\hat{J}_{3D})$};   
  \end{tikzpicture}
   \vspace{-3em}
   \caption{\textbf{Overall architecture}. Given a broadcast video, we begin by extracting player tracklets, denoted as $\mathcal{T} \in \{\mathcal{T}_1, \mathcal{T}_2, ..., \mathcal{T}_n\}$. Each tracklet $\mathcal{T}_k$ consists of a sequence of frames $\mathbf{F}_i$ where $\mathbf{F}_i \in \mathbb{R}^{H \times W \times3}$ for $N$ frames. These tracklets are then processed through a Temporal Convolutional Network (TCN), which implicitly decouples player actions and identifies the tracklet of the pitcher, called $\mathcal{T}_p$. Subsequently, $\mathcal{T}_p$ undergoes encoding via an encoder (\textbf{E}) to derive pseudo-depth information for each frame. The frames, along with their corresponding pseudo-depth data, are fed into a 3D modeling technique (D2A-HMR 2.0). This framework is responsible for predicting the 3D mesh and 3D joint positions of the pitcher, facilitating detailed analysis of various pitch metrics using the temporal kinematic information processing the 3D joint positions.}
   \label{fig:overview}
\end{figure*}
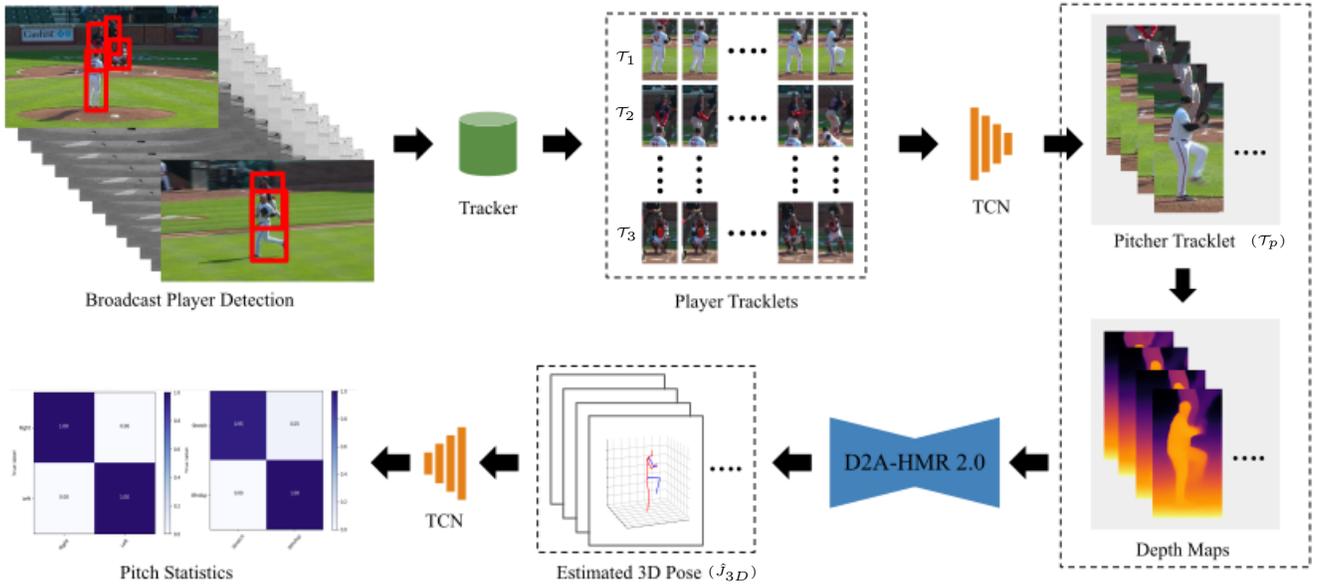
\section{Literature Review} \label{review}

\paragraph{Player Tracking and Identification.} Various approaches for player identification exist, primarily relying on either facial features or jersey numbers. In works such as \cite{face_id1, face_id2}, facial recognition is employed to label players based on detected face regions. Conversely, jersey number recognition, as seen in \cite{mmjersey, DeepPlayer, liu2022deep, Gerke2015SoccerJN, vats_tracking}, is a prevalent method for player identification. Vats \etal \cite{vats_tracking} recently introduced a comprehensive offline tracking framework for ice hockey, employing 1D convolutions for team and jersey number identification. Sentioscope \cite{sentioscope} tracks player interactions using a dual camera setup and model field particles on a calibrated soccer field plane. DeepPlayer \cite{DeepPlayer} proposes a multicamera player identification system integrating jersey number patterns, team classification, and pose-guided partial features. Works such as \cite{CHAN2021113891, mmjersey} incorporate end-to-end trainable spatio-temporal networks for identifying jersey numbers in ice hockey and soccer. Additionally, \cite{pose_id1, pose_id2, DeepPlayer} utilize convolutions to extract features and exploit information from the pose of the players to determine the numbers of the jersey.

Existing player identification approaches rely heavily on distinctive features (such as clothing, jersey number, and facial features). However, these features are often unreliable due to clothing variations, occlusions, and varying camera angles. To address these challenges, we propose a novel approach that decouples player actions from individual tracklets. This approach shifts the focus from specific player features to the action itself, enabling robust and accurate player identification even when traditional features fail. 

\paragraph{Player Action Recognition.} Deep learning has emerged as a powerful tool for action recognition, offering promising results. The use of 3D convolutions has demonstrated effectiveness in capturing crucial spatio-temporal information from video data \cite{li-ar, PoTion, angela-qr, Cai2018TemporalHA}. However, these methods often suffer from a high number of parameters, making them susceptible to overfitting on smaller datasets. To address this limitation, Li \etal \cite{li-ar} introduced a spatio-temporal attention network, enabling identification of the key video frames and spatially focus on those frames. Similarly, works including \cite{Cai2018TemporalHA, PoTion, angela-qr, yao-ar} leveraged the pose features from each frame of the sequence and enable effective action recognition without introducing parameter overhead. Yao \etal \cite{yao-ar} coupled pose and action by formulating pose as an optimization on a set of action-specific manifolds. Cai \etal \cite{Cai2018TemporalHA} utilize a two-stage architecture that extract pose information and temporal information using optical flow technique before combining them. STAR-Transformer \cite{star-trans} fused video and skeletal data using a transformer architecture with a special cross-attention mechanism. SVFormer \cite{svformer} introduced a semi-supervised learning approach by incorporating a novel data augmentation technique called Tube TokenMix, specifically designed to improve video understanding.

While recent advancements in player action recognition have been impressive, a gap remains in fully utilizing the rich numerical data embedded within gameplay videos. Integrating this statistical information alongside visual cues holds significant promise for enhancing recognition accuracy and robustness. By incorporating these game statistics, the models can potentially learn more nuanced patterns in player behavior and become less susceptible to outliers.

\paragraph{Baseball Pitch Statistics.} Previous research has mostly focused on estimating the pitch statistics from existing baseball data collection database such as the PITCHf/x system \cite{bock2015pitch, sidle2018pred, woodward2014decision}. Works such as \cite{bock2015pitch} leveraged these prior game statistics to classify pitch types using Support Vector Machines (SVM) with linear kernal functions, and \cite{sidle2018pred} utilized Linear Discriminant Analysis (LDA), decision trees, and SVM to find the best apt model to classify pitch types. Hickey \etal \cite{hickey2020dissecting} aimed to improve the interpretability alongside accuracy of classification models used for pitch prediction. Recently, Manzi \etal \cite{manzi2024pitch} proposed a descriptive laboratory study in the setting of 3D motion-capture to classify pitch throws by analyzing pitcher kinematics. Similarly, Oyama \etal \cite{oyama2017} used motion capture systems to validate the pitching motion of the baseball by comparing with the calculated angles. Chen \etal \cite{Chen2011} proposed a network which can recognize hand pitching style (overhand, three-quarter, etc.) by extracting the human body segment and a descriptor representation using star skeletons.
\section{Methodology} \label{method}

The overview of the proposed system, PitcherNet is presented in Figure \ref{fig:overview}. The system is divided into three components: (1) \textbf{Pitcher Tracking and Identification}, involving the initial detection of all players, subsequent tracking of detected players with the assignment of unique labels to each tracklet, and the \textit{decoupling of actions} from the inferred poses of the players in each sequence in the tracklet to facilitate player classification; (2) \textbf{3D Human Modeling} utlizes a 3D human model prior \cite{smpl} to estimate the pose of the player guided by masked modeling, distribution learning, and silhouette masks; and (3) \textbf{Pitch Statistics} leveraging TCN and kinematic-driven heuristics to reliably capture various pitch metrics. This section provides a comprehensive exploration of each component, elucidating the underlying design choices that aim to improve the performance of existing techniques in the context of the proposed system.

\subsection{Pitcher Tracking and Identification} \label{subsec:identification}

Accurate tracking and identification of players are fundamental for effective action recognition and analysis in sports scenarios. As highlighted in the literature, the challenges associated with simultaneous tracking and classification based on features are the compromise in reliability of obtaining the desired tracklet. Thus, our objective is to \textit{decouple the action from kinematics obtained from sequences of the tracklets} to acquire the desired tracklet ID for subsequent downstream tasks.

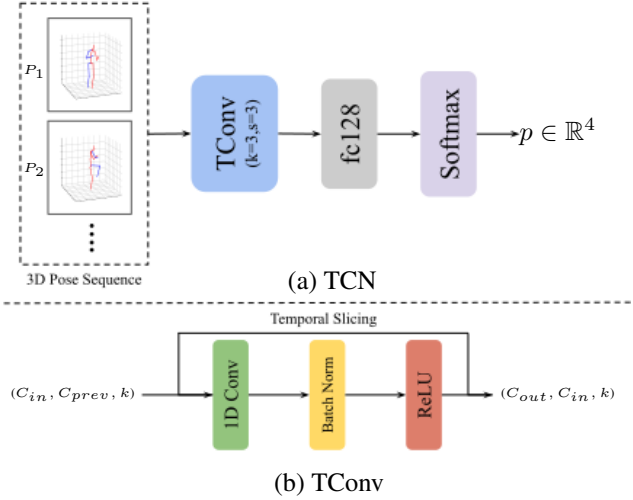
\begin{figure}
{\centering
  \begin{tikzpicture}
    \node at (0,0) {\includegraphics[width=\linewidth]{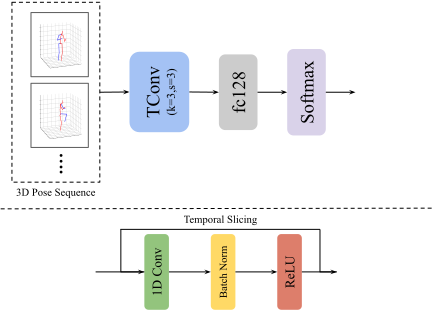}};
    \node[font=\tiny] at (-3.76,2.3) {${P}_1$};
    \node[font=\tiny] at (-3.76,1) {${P}_2$};
    \node[font=\normalsize] at (3.2,1.5) {$p \in \mathbb{R}^{4} $};
    \node[font=\tiny] at (-3.25,-2.0) {($C_{in}, C_{prev}, k$)};
    \node[font=\tiny] at (3.2,-2.0) {($C_{out}, C_{in}, k$)};
    \node[font=\normalsize] at (0.2,-0.5) {$\text{(a) TCN}$};
    \node[font=\normalsize] at (0.2,-3.2) {$\text{(b) TConv }$};
  \end{tikzpicture}
  \vspace{-20px}
  \caption{ \textbf{Temporal Convolutional Network.} \textbf{(a)} Overview of the proposed TCN for the player identification task, where $\text{fc}$ denotes fully connected layers and $p$ refers to the model's output. \textbf{(b)} Architecture of the TConv block used in the TCN, where $C_{in}$, $C_{out}$ and $C_{prev}$ denotes the input, output and previous channels, respectively and $k$ denotes the kernel size.}
\label{fig:tcn}
}
\end{figure}

Initially, tracklets are generated using the methodology proposed in SORT \cite{sort}, which utilize YOLOX \cite{yolox} detections. Each tracklet is assigned a unique identifier along with the 3D pseudo-pose from MHFormer \cite{mhformer}. MHFormer is chosen due to its lightweight nature which allows efficient processing of the tracklets to estimate player roles. Subsequently, from the pseudo-pose, we decouple player actions by classifying each tracklet into the player's role (pitcher, batter, or others). Given that pitchers are the primary focus of our investigation, we identify sequences within tracklets where pitching actions occur. To accomplish this, we employ a Temporal Convolutional Network (TCN) architecture designed to decouple various actions within each tracklet, specifically isolating the pitching action of interest. The TCN architecture, described in Figure \ref{fig:tcn}, eliminates the dependence on the characteristics of specific players for classification, providing a more robust solution to identify the target player in dynamic sports scenarios.

The TCN architecture utilizes a series of five TConv layers which encompass a dilated 1D convolutional layer with progressively increasing dilation rates, followed by batch normalization and ReLU activation in each layer. The network ingests a 4D tensor representing pose sequences $\mathbb{P} = \{P_i:P \in \mathbb{R}^{K \times C}\}^N_{i=0}$, where each dimension corresponds to batch size ($B$), temporal sequence length ($N$), number of joint positions ($K$), and 3D player coordinates ($C$). This progressive dilation allows the TCN to capture long-range temporal dependencies crucial for understanding complex motion patterns, while dropout layers and batch normalization enhance the model's generalizability. In addition, skip connections are utilized, allowing the model to directly access information from the original input at a deeper stage in the network. This address the problem of vanishing gradients and improve the flow of information throughout the TCN architecture.

\subsection{3D Human Modeling} \label{sec:3dpose}

Estimating the pose of the pitcher is crucial for effective pitch analysis of the players from live broadcast video. The input to the 3D human modeling technique is the player of interest tracklet from the pitcher tracking and identification component (Section \ref{subsec:identification}). To enhance the reliability of pose estimation in challenging, real-world scenarios, we have advanced the D2A-HMR technique introduced in \cite{d2ahmr} specifically focusing on our input data conditions.   

D2A-HMR focuses on learning the underlying output distribution to minimize the distribution gap. The method takes an input image and pseudo-depth maps, utilizing a transformer encoder that incorporates cross- and self-attention mechanisms along with learnable gate fusions to produce the final output token. Subsequently, a decoder is employed to predict the person's silhouette, guiding the overall structure of the player in the input image. Additionally, a regression head is utilized to obtain the mesh vertices. In this work, we have introduced several design enhancements to augment the D2A-HMR model, and these modifications will be referred to as the D2A-HMR 2.0 modeling technique.

D2A-HMR 2.0 leverages a depth encoder called Depth Anything \cite{depthanything} to extract pseudo-depth, which utilizes the DINOv2 encoder \cite{dinov2} and the DPT decoder \cite{dpt}. In addition to regression of the mesh vertices as output, we extract the 3D joint coordinates ($J_{3D}$). Following \cite{hmr, spin, pose2mesh}, the mesh vertices are further regressed to find the 3D regressed joint coordinates ($J_{3D}^r$) using a predefined regression matrix $G \in \mathbb{R}^{K \times M}$. Here $K$ and $M$ correspond to the number of joint positions and the number of vertices. Then, the final 3D pose ($\hat{J}_{3D}$) of the person in the input image is formulated as shown in Equation \eqref{eq:pose}.

\begin{equation}
    \hat{J}_{3D} = \omega_1 J_{3D} + (\omega_2 V_{3D})G  = \omega_1 J_{3D} + \omega_2 J_{3D}^r
    \label{eq:pose}
\end{equation}

where $\omega_1$ and $\omega_2$ are weights for the joint distribution. $V_{3D}$ denotes the vertices of the output mesh. 

\begin{figure}[t]
{\centering
  \begin{tikzpicture}
    \node at (0,0) {\includegraphics[width=\linewidth]{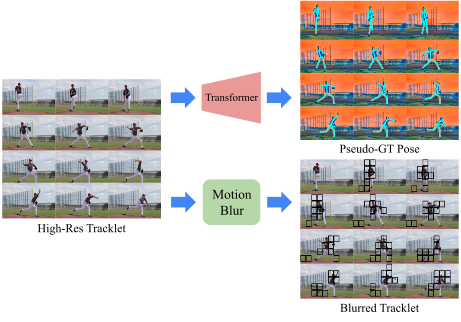}};
  \end{tikzpicture}
  \vspace{-25px}
  \caption{\textbf{Data Augmentation Technique.} Pseudo-ground truth pose is collected using a Transformer model for improved generalizability of the pose estimation model.}
\label{fig:teacher}
}
\end{figure}

To further enhance the efficacy of D2A-HMR 2.0, we integrate a substantial volume of unlabeled data sourced from the Internet to facilitate robust generalization in dynamic real-world scenarios, as illustrated in Figure \ref{fig:teacher}. This augmentation involved the implementation of a transformer network with pretrained weights initialized using MHFormer \cite{mhformer}. Specifically, we select high-resolution practice videos from the Internet and introduce motion blur artifacts. Prior to inducing blur, we use a transformer network to predict its corresponding 3D poses, leveraging its superior performance with high-resolution data. 

\subsection{Pitch Statistics} \label{method:ar}

The pitch type, a complex interplay of various pitch statistics, ultimately determines the pitch delivered. This work focuses on estimating crucial kinematically derived pitch statistics such as pitch position, release point, release extension, pitch velocity, and handedness from the 3D pose data obtained using the D2A-HMR 2.0 model. While factors such as break and spin rate also influence pitch action \cite{nasu2021impact}, this work focuses on these core kinematic statistics mentioned above. By analyzing these statistics, we gain valuable insight into the mechanics of pitch delivery. These pitch statistics combined with kinematic motion data will contribute significantly in the prediction of complex pitch actions. The 3D kinematic information is fed as input to the pitch statistics component to estimate the different pitch statistics, including pitch position, release point, pitch velocity, release extension, and handedness. 

\subsubsection{Pitch Position} Pitchers utilize two legal pitch position styles: the windup, a full-body motion maximizing power, and the set/stretch, a quicker, more compact motion sacrificing some velocity for faster delivery. Mastering these positions allows pitchers to deceive batters by disrupting their timing and pitch recognition \cite{scarborough2021association}. This work uses a TCN backbone, identical to the player identification network, for pitch position classification using a sigmoid activation function. Each video tracklet is fed into the TCN with a sequence length of 100 frames for classification.

\subsubsection{Handedness} Accurate determination of the pitcher's handedness is critical for effective pitch analysis. By isolating the throwing hand within each video frame, we can tailor feature extraction to the pitcher's specific mechanics. This work utilizes a TCN to estimate handedness. While the TCN demonstrates effectiveness, simpler methods based on hand appearance analysis might also be suitable for handedness classification. Regardless of the chosen technique, identifying handedness allows the analysis pipeline to account for the pitcher's mechanics, leading to improved feature extraction and ultimately, more accurate pitch statistics.

\subsubsection{Release Point} \label{sub:releasept} The release point, defined as the specific location where the pitcher releases the ball from their hand toward the batter, plays a crucial role in determining both the pitch velocity and the release extension \cite{whiteside2016ball}. It also plays a crucial factor in deciphering tunneled pitchers.

\begin{figure}[H]
{\centering
\includegraphics[width=\linewidth]{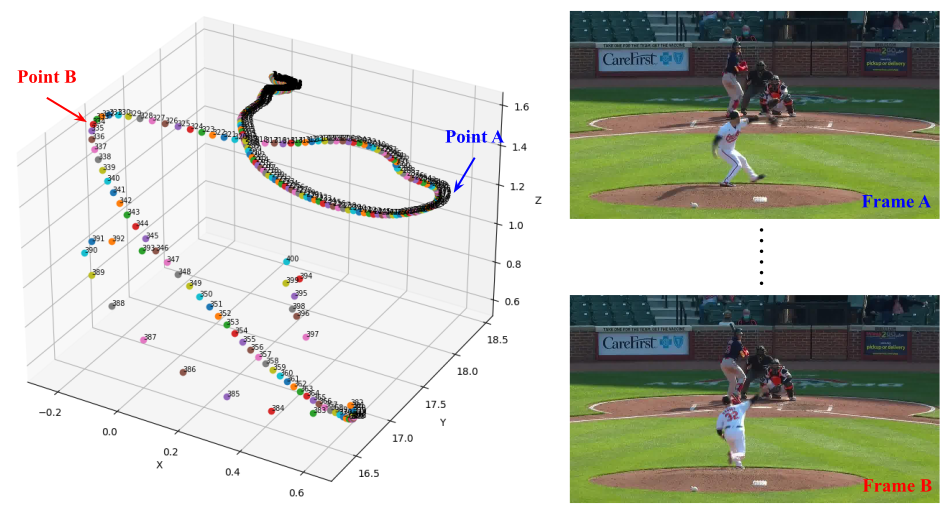}
\caption{\textbf{Trajectory of the right wrist joint in 3D space.} Illustration of two frames which correspond to the points (A and B) marked in the trajectory plot that determines the release point.}
\label{fig:release-point}
}
\end{figure}

As illustrated in Figure \ref{fig:release-point}, we use the wrist kinematics of the pitcher in the x-plane (lateral movement) to identify the release point. Here, we determine the extreme coordinates in the x-plane to establish the maximum and minimum limits of wrist movement during the throwing motion. The limit point, Point A corresponds to the final cocking phase which refers to the point of maximum external rotation of the throwing shoulder from the glove,  while the limit point Point B, represents the end of the acceleration phase with the ball release and start of the follow-through phase \cite{murray2001effects}. We hypothesis that the ball release point will be one amongst $n$ frames windowed on point B with the peak pitch velocity.  

\subsubsection{Pitch Velocity} Pitch velocity, measured in miles per hour, reflects the ball's speed upon leaving the pitcher's hand \cite{Fortenbaugh2009}. This study proposes a method to estimate pitch velocity by analyzing changes in the throwing hand's wrist position at the release point identified using the estimated 3D pose data. Equation \eqref{eq:pitch_vel} calculates the angular velocity of the wrist at release based on the change in arctangent of consecutive wrist coordinates ($w_x$, $w_y$) before (frame $r-1$) and at (frame $r$) the release point. This angular velocity is then converted to an approximate pitch velocity ($v_p$) by multiplying it by the lever arm length ($l$) between the wrist and the elbow joint.

\begin{equation}
    v_p = \omega \times l =  \{(atan(w_y^{r}, w_x^{r}) - atan(w_y^{r-1}, w_x^{r-1})) \times {T}\} \times {l}
    \label{eq:pitch_vel}
\end{equation}

As mentioned above in Section \ref{sub:releasept}, we estimate the ball release point by finding the maximum velocity in a window around Point B using Equation \eqref{eq:pitch_vel} which will compute the pitch velocity.  

\subsubsection{Release Extension} Release extension, in baseball, refers to the distance a pitcher creates between the pitching mound and the release point of the ball towards the batter. The release extension helps to differentiate pitch types, as fastballs typically involve greater extension compared to breaking balls. It essentially describes how much closer the pitcher gets to the home plate at the moment of release compared to where they started their throwing motion. The release extension is mathematically indicated as shown in Equation \eqref{eq:extension}.

\begin{equation}
    \text{Extension} = \sqrt{(w_x-a_x)^2 + (w_y-a_y)^2 + (w_z-a_z)^2}
    \label{eq:extension}
\end{equation}

Here, in Equation \eqref{eq:extension}, $a$ refers to the position of the ankle joint of the pitching leg. The pitching ankle joint is chosen since it remains planted on the mound during the pitch set and the initial part of the delivery.  

\subsection{Loss Functions} \label{subsec:loss}

Most pitchers tend to be right-handed. Therefore, there is an imbalance in the class, especially in the handedness data of the pitchers. We use focal loss as a loss function to address the issue of class imbalance \cite{focal}. More weight is given to minority classes while training using a gamma tuning parameter (set to 2 initially). The loss function used for the estimation of pitch position and handedness is denoted as shown in Equation \eqref{eq:focal_loss}.

\begin{equation}
    L(p_t) = -\alpha_t * (1 - p_t)^\gamma * \log(p_t) \label{eq:focal_loss}
\end{equation}

where $\alpha_t$ and $\gamma$ are the balancing parameter and sampling focus parameter, respectively. $p_t$ denotes the predicted probability of the true class. The D2A-HMR human model is trained using the objective mentioned in Equation \eqref{eq:loss}.

\begin{equation}
    \begin{aligned}
        \mathcal{L}_{model} &= \lambda_{RLE} \mathcal{L}_{RLE} + \lambda_{SMPL} \mathcal{L}_{SMPL} + \lambda_{3D} \mathcal{L}_{3D} \\
        &\quad + \lambda_{2D} \mathcal{L}_{2D} + \lambda_{silh} \mathcal{L}_{silh}
    \end{aligned}
    \label{eq:loss}
\end{equation}

where $\mathcal{L}_{RLE}$, $\mathcal{L}_{SMPL}$, $\mathcal{L}_{3D}$, $\mathcal{L}_{2D}$ and $\mathcal{L}_{silh}$ correspond to residual likelihood loss, SMPL vertex loss, regressed 3D loss, reprojected 2D loss and silhouette loss. All $\lambda$ correspond to the weights assigned to distribute the importance of each objective. We incorporated an additional loss function into our D2A-HMR 2.0 as shown in Equation \eqref{eq:new_loss}. 

\begin{equation}
    \hat{\mathcal{L}}_{model} = \mathcal{L}_{model} + \lambda_{3D}^r \mathcal{L}_{3D}^r
    \label{eq:new_loss}
\end{equation}

Here, $\mathcal{L}_{3D}^r$ corresponds to the 3D output joints of the regression head of the D2A-HMR model. 
\section{Experimentation} \label{exp}

\paragraph{Implementation Details.} The training process is conducted on three NVIDIA A6000 GPUs with 48GB RAM. Adam optimizer with a batch size of 48 with 500 epochs is used to train the 3D human model. A learning rate of $10^{-4}$ with betas of 0.9 and 0.99 is used for optimization. The TCN model for handedness estimation and pitch position estimation is trained for 50 and 100 epochs, respectively using one of the three GPUs. The TCN model trained for player identification was trained for 200 epochs using two GPUs with AdamW optimizer with a learning rate of $10^{-2}$.

\paragraph{Dataset.} We utilize the MLBPitchDB dataset \cite{d2ahmr} to evaluate our system performance. This dataset is specifically designed for baseball sports analysis and encompasses various data points including player details, 3D pose estimations, actions, and play statistics for all players within the camera's field of view (FoV). Some available ground truth play statistics from the dataset includes pitch velocity, pitch extension, release point of the ball, spin rate, handedness, vertical and horizontal break. Before training, the dataset is pre-processed following the techniques described in \cite{mitigatingblur}, which involved player detection, data synchronization, and camera reprojection. To train the 3D human model used for estimating the pose from input images, we utilize the Human 3.6M \cite{h36m} and 3DPW \cite{3dpw} datasets. In the case of Human3.6M, we trained our D2A-HMR model 2.0 on subjects S1, S5, S6, S7, and S8 and conducted testing on S9 and S11. These configurations were aligned with the common training and evaluation settings within the domain \cite{pose2mesh, d2ahmr}.

\paragraph{Metrics.} We employ several metrics to assess the performance of different components of our system. For the 3D human pose estimation task, we utilize two primary metrics: the mean per joint position error (mPJPE) and the procrustes-aligned mean per joint position error (PA-mPJPE). These metrics quantify the average distance between the predicted and ground truth 3D joint locations, with PA-mPJPE accounting for global pose variations. For player tracking and identification, we evaluate the system's accuracy performance. Finally, pitch statistics performance is assessed using a standard accuracy metric with different classification margins. 

\subsection{Pitcher Identification}

The impact of the pitcher identification task is compared with two temporal networks (LSTM, transformer with only self-attention blocks) in Table \ref{tab:class-role}. Simple baseline temporal networks were used for comparison to validate the effectiveness of pose-based role classification. Since these are tasks that use distinct player kinematics, complex networks were not needed.

\begin{table}[H]
  \centering
  \caption{Comparison of our model with baseline temporal networks on MLBPitchDB dataset \cite{d2ahmr}.}
  \begin{tabular}{@{}lc@{}}
    \toprule
     & Test Accuracy $\uparrow$   \\
    \midrule
    LSTM & 85.55 \\
    Transformer & 91.11 \\
    \midrule
    \textbf{Ours} & \textbf{96.66} \\
    \bottomrule
  \end{tabular}
  \label{tab:class-role}
\end{table}

Our approach achieves superior test accuracy compared to both LSTMs and transformers with self-attention blocks, as shown in Table \ref{tab:class-role}. Specifically, we observe an improvement of 11.11\% and 5.55\% in accuracy relative to LSTMs and transformers, respectively.

\subsection{3D Human Modeling}

\paragraph{Depth Encoder.} Experimentation with different depth encoders including AdaBin \cite{adabins}, ZoeDepth \cite{zoedepth}, DINOv1 \cite{dinov1}, DINOv2 \cite{dinov2} and Depth Anything \cite{depthanything} is done in Table \ref{tab:depth}. The D2A-HMR model proposed in \cite{d2ahmr} utilizes DINOv2 \cite{dinov2} as its depth encoder for human body modeling. 

\begin{table}[H]
  \centering
  \caption{Impact on different depth encoders for D2A-HMR evaluated on 3DPW dataset.}
  \begin{tabular}{@{}lcc@{}}
    \toprule
     & mPJPE $\downarrow$ & PA-mPJPE $\downarrow$ \\
    \midrule
    Bhat \etal \cite{adabins} & 90.3 & 55.4 \\
    Bhat \etal \cite{zoedepth} & 87.8 & 53.3 \\
    Caron \etal \cite{dinov1} & 83.1 & 50.6 \\
    Oquab \etal \cite{dinov2} & 80.5 & 48.4 \\
    \midrule
    \textbf{Yang \etal \cite{depthanything}} & \textbf{78.7} & \textbf{46.9} \\
    \bottomrule
  \end{tabular}
  \label{tab:depth}
\end{table}

Our findings demonstrate that employing Depth Anything \cite{depthanything} as the depth encoder to generate pseudo-depth leads to improvements of 1.8mm and 1.5mm in mPJPE and PA-mPJPE, respectively. This can be attributed to using a significantly larger dataset of unlabeled images allowing the model to learn more comprehensive visual representations. Refer to the Appendix for a qualitative comparison of the various monocular depth estimation techniques on the MLBPitchDB dataset. 

\begin{figure}[t]
{\centering
  \begin{tikzpicture}
    \node at (0,0) {\includegraphics[width=\linewidth]{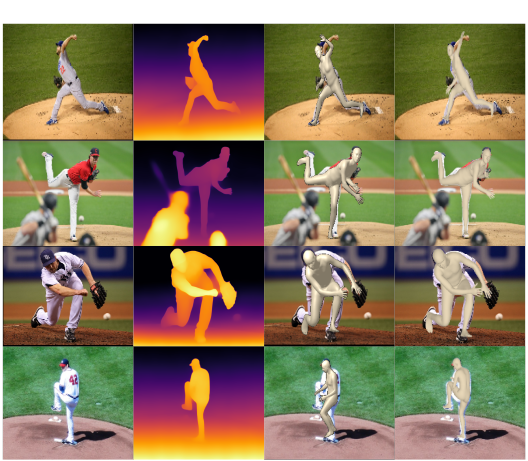}};
    \node[font=\normalsize] at (-3.1,3.5) {$\text{Input}$};
    \node[font=\normalsize] at (-1.1,3.5) {$\text{Pseudo-Depth}$};
    \node[font=\normalsize] at (1,3.5) {$\text{D2A-HMR}$};
    \node[font=\normalsize] at (3,3.5) {$\text{Ours}$};
  \end{tikzpicture}
  \vspace{-20px}
  \caption{\textbf{Qualitative results.} Qualitative comparison of the pitcher's mesh alignment with the input image using our D2A-HMR 2.0 model.}
\label{fig:d2a-hmr}
}
\end{figure}
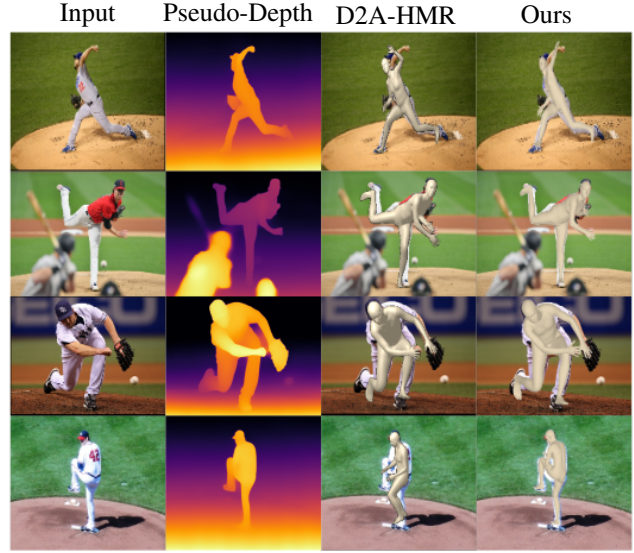

\paragraph{Regression Heads.} We evaluate the performance of two distinct regressor head architectures within our model. The first design directly regresses the vertex coordinates of the transformer output tokens. On the contrary, the second approach predicts both the vertices and the 3D pose of the players. 

\begin{table}[H]
  \centering
  \caption{Ablation study on different regressor heads for D2A-HMR evaluated on 3DPW dataset.}
  \begin{tabular}{@{}lcc@{}}
    \toprule
     & mPJPE $\downarrow$ & PA-mPJPE $\downarrow$ \\
    \midrule
    w/ vertex only & 80.5 & 48.4 \\
    w/ vertex+joints & \textbf{78.7} & \textbf{46.9} \\
    \bottomrule
  \end{tabular}
  \label{tab:regressor}
\end{table}

\begin{table*}
    \caption{Performance of our pitch statistics module on different pitch metrics including pitch handedness, pitch position, release point, pitch velocity, and release extension in the test dataset compared against baseline temporal networks.}
    \centering
    
    \begin{subtable}{0.4\linewidth}
      \centering
      \caption{Handedness}
      \begin{tabular}{@{}lccc@{}}
        \toprule
         & Accuracy $\uparrow$ & F1 Score  $\uparrow$ & Precision $\uparrow$ \\
        \midrule
        LSTM & 85.0 & 85.7 & 90.0 \\
        \textbf{Ours (TCN)} & \textbf{100.0} & \textbf{100.0} & \textbf{100.0} \\
        \bottomrule
      \end{tabular}
      \label{tab:handedness}
    \end{subtable}
    \hspace{25pt}
    \begin{subtable}{0.4\linewidth}
      \centering
      \caption{Pitch Position}
      \begin{tabular}{@{}lccc@{}}
        \toprule
         & Accuracy $\uparrow$ & F1 Score  $\uparrow$ & Precision $\uparrow$ \\
        \midrule
        LSTM & 81.3 & 82.5 & 85.0 \\
        \textbf{Ours (TCN)} & \textbf{97.5} & \textbf{97.4} & \textbf{95.0} \\
        \bottomrule
      \end{tabular}
      \label{tab:pitcherset}
    \end{subtable}
    \vspace{5pt}
    \vspace{0pt}
    \begin{subtable}{0.3\linewidth}
      \centering
      \caption{Release Point}
      \begin{tabular}{@{}lccc@{}}
          \toprule
         & $A_{1}$ $\uparrow$ & $A_{2}$ $\uparrow$ & $A_{5}$ $\uparrow$\\ 
        \midrule
        LSTM & 31.3 & 46.4 & 63.5 \\
        TCN & 43.4 & 51.5 & 77.6 \\
        \textbf{Ours} & \textbf{80.8} & \textbf{85.8} & \textbf{97.9} \\
        \bottomrule
      \end{tabular}
      \label{tab:release-point}      
    \end{subtable}
    \hspace{5pt}
    \vspace{0pt}
    \begin{subtable}{0.3\linewidth}
      \centering
      \caption{Pitch Velocity}
      \begin{tabular}{@{}lccc@{}}
        \toprule
         & $A_{1\%}$ $\uparrow$ & $A_{2\%}$ $\uparrow$ & $A_{5\%}$ $\uparrow$\\ 
        \midrule
        LSTM & 5.1 & 13.1 & 22.2 \\
        TCN & 10.1 & 18.1 & 48.4 \\
        \textbf{Ours} & \textbf{43.4} & \textbf{68.6} & \textbf{94.9} \\
        \bottomrule
      \end{tabular}
      \label{tab:vel}
    \end{subtable}
    \hspace{20pt}
    \vspace{0pt}
    \begin{subtable}{0.3\linewidth}
      \centering
      \caption{Release Extension}
      \begin{tabular}{@{}lccc@{}}
        \toprule
         & $A_{5\%}$ $\uparrow$ & $A_{8\%}$ $\uparrow$ & $A_{10\%}$ $\uparrow$\\ 
        \midrule
        LSTM & 4.0 & 7.1 & 11.1 \\
        TCN & 14.1 & 19.1 & 25.2 \\
        \textbf{Ours} & \textbf{24.2} & \textbf{31.3} & \textbf{37.3} \\
        \bottomrule
      \end{tabular}
      \label{tab:extension}
    \end{subtable}
      
    \label{tab:pitch_statistics}
\end{table*} 

As shown in Table \ref{tab:regressor}, our model demonstrates superior performance when incorporating both the player's vertices and 3D joints during the regression process. This is likely due to the additional information provided by the joints, which helps the model refine the predicted 3D pose and achieve more accurate alignment. Furthermore, the model optimizes the output 3D joints by minimizing the difference between them and the ground truth 3D poses, further contributing to the overall improvement in performance. Figure \ref{fig:d2a-hmr} demonstrates the superior alignment ability of D2A-HMR 2.0 from a given input image when compared with existing state-of-the-art HMR techniques. 

\paragraph{Pseudo-GT Data.} The impact of using additional pseudo-ground truth pose data on the training process of the D2A-HMR 2.0 model is presented in Table \ref{tab:gen}. Specifically, pseudo 2D and 3D pose data obtained from HRNet \cite{hrnet} and MHFormer \cite{mhformer}, respectively, are used as ground truth to train the HMR model.
 
\begin{table}[H]
  \centering
  \caption{Ablation study on utilizing Pseudo-GT data.}
  \begin{tabular}{@{}lcc@{}}
    \toprule
     & mPJPE $\downarrow$ & PA-mPJPE $\downarrow$ \\
    \midrule
    w/o Pseudo-GT data & 79.1 & 47.4 \\
    w/ Pseudo-GT data & \textbf{78.7} & \textbf{46.9} \\
    \bottomrule
  \end{tabular}
  \label{tab:gen}
\end{table}

Our D2A-HMR 2.0 model demonstrates an improvement in the performance of recovering the 3D mesh by incorporating additional pseudo-ground truth data from the internet, as shown in Table \ref{tab:gen}. 

\subsection{Pitch Statistics} \label{subsec:ar}

Table \ref{tab:pitch_statistics} shows the pitch statistics from the broadcast videos. TCN with five TConv blocks is utilized to predict handedness and pitch position from the kinematic motion sequence of the pitcher. The ball release point is extracted using heuristics from the trajectory of the wrist position of the pitcher and validated using the ground truth release point information from the MLBPitchDB dataset. The pitch velocity and release extension are then computed using mathematical functions that utilize ball release point and kinematic pose information.  

Table \ref{tab:handedness} demonstrates that the TCN model achieves perfect accuracy (100\%) in classifying the handedness, without misclassifications for right-handed or left-handed pitchers. The model demonstrates impressive performance in classifying pitch position, correctly identifying 95\% of the stretch deliveries and 100\% of windup deliveries as shown in Table \ref{tab:pitcherset}. The misclassification rate is low, with only 5\% of stretch deliveries misclassified as windup, and no misclassification observed for windup deliveries. 

The validity of the approach adapted to estimate the release point is examined and compared with alternative methods in Table \ref{tab:release-point}. The table shows that directly inferring the release point from a temporal network tends to perform poorly. ${A}_x$ denotes the accuracy with $x$ the number of frames as a margin in the table. Table \ref{tab:vel} shows that our method estimates pitch velocity with superior performance compared to existing temporal networks. $A_{x\%}$ in Table \ref{tab:vel} denotes the accuracy with $x\%$ as the margin. Finally, Table \ref{tab:extension} presents the superior performance to estimate the release extension. It can be further improved by studying the pitching stride length, which is the distance covered between the spot where one foot hits the ground and the next time the same foot hits the ground again \cite{Fortenbaugh2009}. Our method directly utilizes the release point and 3D pose information to calculate the extension, achieving accurate results compared to the baseline networks.
\section{Conclusion} \label{conc}

We introduce PitcherNet, an end-to-end deep learning system for kinematic-driven pitch analysis in baseball sports through robust 3D human modeling from broadcast videos. By overcoming challenges such as motion blur in low-resolution feeds, PitcherNet accurately identifies a range of pitch statistics, including pitch position, release point, pitch velocity, pitcher handedness, and release extension. This empowers players, coaches, and fans to gain deeper insights into the technical nuances of pitching and unlock strategic advantages. Additionally, decoupling action from tracklets paves the way for reliable player identification, which holds potential for sports analytics and performance evaluation. 

Future work includes extending the framework to analyze further crucial pitch statistics, particularly those involving the pitcher's grip, and delving deeper into player mechanics to optimize posture, reduce injury risk, and further enhance performance.

\section*{Acknowledgments}

This work is supported by the Baltimore Orioles, Major League Baseball through the Mitacs Accelerate Program. We are grateful for their support and contributions, which have significantly enriched the research and its findings. 
{
    \small
    \bibliographystyle{ieeenat_fullname}
    \bibliography{main}
}

\clearpage
\setcounter{page}{1}
\maketitlesupplementary
\appendix

The supplementary material is organized into the following sections:

\begin{enumerate}
    \item Section \ref{app:d2ahmr}: Architecture of D2A-HMR 3D human modeling technique.
    \item Section \ref{app:pitch_id}: Comparison of the proposed pitcher identification network with jersey number techniques.
    \item Section \ref{app:qual}: Qualitative comparison of the PitcherNet system and various components including the depth encoders and D2A-HMR 2.0.
    \item Section \ref{app:limitations}: Limitations of the proposed system.
    
\end{enumerate}

\section{D2A-HMR Architecture} \label{app:d2ahmr}

In this section, we explain the D2A-HMR architecture proposed in \cite{d2ahmr} in detail. D2A-HMR leverages a transformer-based architecture by incorporating scene-depth information, which is crucial to resolving the ambiguities inherent in single-image data. By jointly learning the distribution of human body shapes and scene-depth, D2A-HMR aims to produce robust 3D human mesh reconstructions, especially for scenarios with unseen data variations.

\begin{algorithm}
\caption{Distribution and Depth Aware Human Mesh Recovery}
\begin{algorithmic}[1]
    \State \textbf{Input:} Image ($\textbf{I}$)
    
    \State \textbf{Initialization:}
    \State $\textbf{\textit{E}(I)} \rightarrow \textbf{D}$
    \State $\textbf{\textit{F}(I, D)}$
    
    \State \textbf{Positional Embedding:}
    \State $P_e (= \omega_1 P_l + \omega_2 P_s) \rightarrow z_{\text{img}}$, $z_{\text{depth}}$

    \State \textbf{Self-Attention (MHSA):}
    \State MHSA($z_{\text{img}}) \rightarrow z'_{\text{img}}$
    \State MHSA($z_{\text{depth}}) \rightarrow z'_{\text{depth}}$
    
    \State \textbf{Cross-Attention (MHCA):}
    \State MHCA($z'_{\text{img}}$, $z'_{\text{depth}}) \rightarrow z_{\text{c}}$
     
    \State \textbf{Learnable Fusion Gates:}
    \State $z$ = $\omega_3 z'_{\text{img}}$ + $\omega_4 z'_{\text{depth}}$ + $(1 - \omega_3 - \omega_4)z_{\text{c}}$

    \State \textbf{Masked Modeling:}
    \State $q_{mask}$ = Mask$(z)$   
    
    \State \textbf{Distribution Matching: }
    \State $\textbf{\textit{R}}$($z$) $\rightarrow \sigma$, $\mu$ 
    \State $\Bar{\mu} = (\mu - \mu_{gt}) / \sigma \rightarrow NF \rightarrow \mathcal{L}_{RLE}$
    
    \State \textbf{Silhouette Decoder:}
    \State $\textbf{I}_{silh}$ = $\textit{\textbf{D}}(z, k, s, p)$
    
    \State \textbf{Output:} 3D mesh vertices, $\mathcal{P} = \textbf{\textit{R}}(z)$, $\mathcal{P} \in \Re^{6890 \times 3}$  
\end{algorithmic}
\label{alg:d2ahmr}
\end{algorithm}

Algorithm \ref{alg:d2ahmr} outlines the core steps of the D2A-HMR model. Initially, a depth encoder $E(I)$ takes an input image ($I$) and generates a depth map ($D$). Concurrently, both $I$ and $D$ are fed as  input to the feature extractor ($F$), followed by hybrid positional encoding $P_e$, yielding tokens $z_{img}$ and $z_{depth}$. These tokens subsequently undergo processing by self-attention and cross-attention modules, resulting in  $z'_{img}$, $z'_{depth}$ and $z_c$ respectively. Fusion gates then merge these outputs into a singular token, $z$. 

To enhance model performance, three refinement modules are employed: masked modeling, distribution modeling, and a silhouette decoder. A log-likelihood residual approach facilitates distribution modeling, enabling the model to learn deviations in the underlying distribution, and consequently generalize more effectively to unseen data. Additionally, masked modeling and a dedicated silhouette decoder refine the mesh shape and feature representation. 

\section{Pitcher Identification} \label{app:pitch_id}

The impact of the pitcher identification task is compared with the classical techniques that use jersey number cues are presented in Table \ref{tab:class-sota}. The classification labels include pitcher, batter, catcher, and player (which includes the fielders and referee). The inputs for the classification task are all the tracklets obtained from the player detection and tracking algorithm, with outputs representing the class for the tracklet detections. 

\begin{table}[H]
  \centering
  \caption{Comparison of our model with state-of-the-art jersey number identification techniques on MLBPitchDB dataset \cite{d2ahmr}.}
  \begin{tabular}{@{}lc@{}}
    \toprule
     & Test Accuracy $\uparrow$   \\
    \midrule
    Gerke \etal \cite{Gerke2015SoccerJN} & 64.47 \\
    Li \etal \cite{STN} & 88.29 \\
    Vats \etal \cite{vats-trans} & 89.46 \\
    Balaji \etal \cite{mmjersey} & 93.68 \\
    Balaji \etal \cite{balaji2024domainguided} & 94.70 \\
    \midrule
    \textbf{Ours} & \textbf{96.82} \\
    \bottomrule
  \end{tabular}
  \label{tab:class-sota}
\end{table}

\begin{figure*}
{\centering
  \begin{tikzpicture}
    \node at (0,0) {\includegraphics[width=\linewidth]{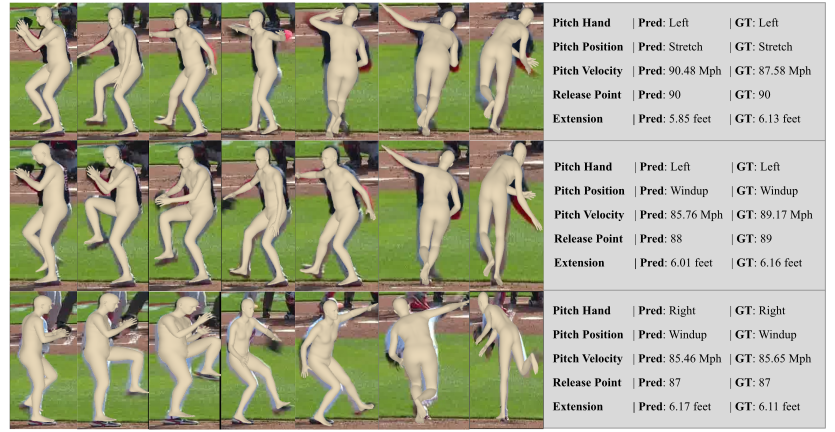}};
  \end{tikzpicture}
  \vspace{-20px}
  \caption{\textbf{Qualitative results.} Performance of the PitcherNet system in capturing various pitch statistics from the player tracklets. Here, \textit{P
  red.} denotes the prediction from the 3D pose information and \textit{GT} denotes the ground truth game data.}
\label{fig:qualitative_sup}
}
\end{figure*}

Table \ref{tab:class-sota} shows that methods that only rely on jersey numbers for player identification underperform on this dataset due to the frequent absence of visible jersey numbers in many video frames. This highlights the importance of decoupling player actions within individual tracklets (sequences of detections associated with a single player) for improved identification accuracy. Therefore, our proposed approach, which incorporates a TCN block to decouple the underlying actions in each tracklet, achieves a significant performance increase of 2.12\% compared to methods solely dependent on jersey numbers.


 


\begin{figure}[H]
{\centering
  \begin{tikzpicture}
    \node at (0,0) {\includegraphics[width=\linewidth]{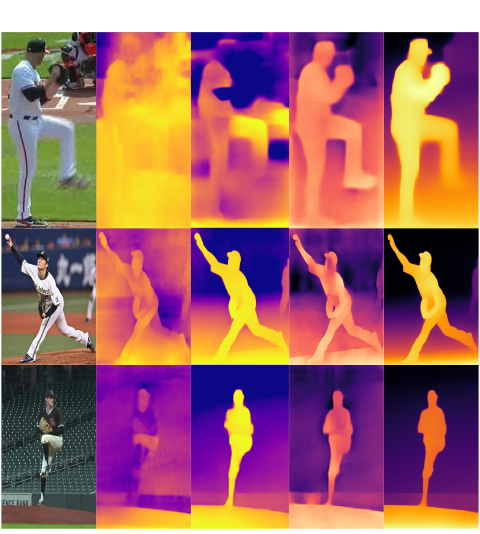}};
    \node[font=\normalsize] at (-3.35,4.4) {$\text{Input}$};
    \node[font=\normalsize] at (-1.65,4.4) {$\text{\cite{adabins}}$};
    \node[font=\normalsize] at (0,4.4) {$\text{\cite{zoedepth}}$};
    \node[font=\normalsize] at (1.65,4.4) {$\text{\cite{dinov2}}$};
    \node[font=\normalsize] at (3.35,4.4) {$\text{\cite{depthanything}}$};
  \end{tikzpicture}
  \vspace{-20px}
  \caption{\textbf{Qualitative results.} Qualitative comparison of the various depth estimation techniques in MLBPitchDB baseball dataset.}
\label{fig:depth}
}
\end{figure}
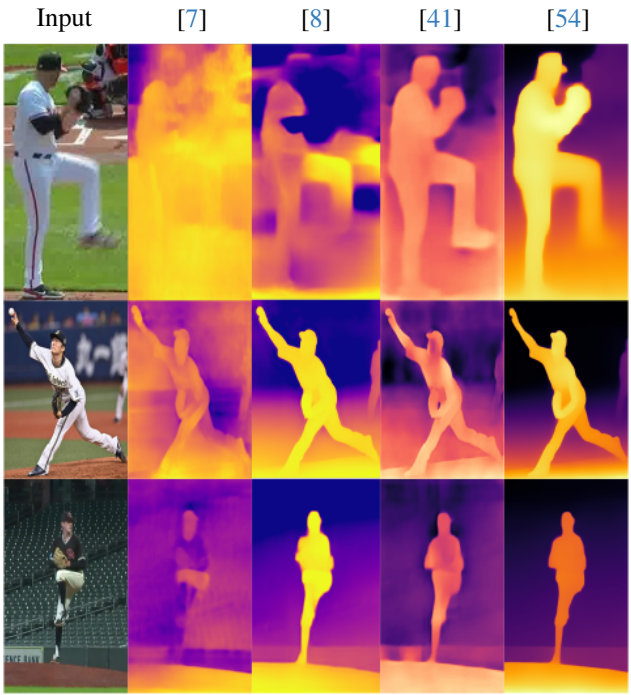

\section{Qualitative Results} \label{app:qual} 

\paragraph{PitcherNet System.} The provided results in Figure \ref{fig:qualitative_sup} highlight the qualitative performance of the PitcherNet system in the MLBPitchDB dataset \cite{d2ahmr}. These visualizations underscore the effectiveness and robustness of our system in achieving accurate alignment with input pitch tracklets.

\paragraph{Depth Encoder.}  Our approach utilizes a monocular depth estimation model as the initial step in the 3D human model generation process. Figure~\ref{fig:depth} qualitatively compares the performance of various techniques, including AdaBin \cite{adabins}, ZoeDepth \cite{zoedepth}, DINOv2 \cite{dinov2}, and Depth Anything \cite{depthanything}. As evident from the figure, Depth Anything \cite{depthanything} exhibits consistently superior depth estimation accuracy compared to the other methods. Consequently, we leverage \cite{depthanything} as the depth encoder within our 3D human model framework.

\begin{figure*}
  \centering
  \includegraphics[width=\linewidth]{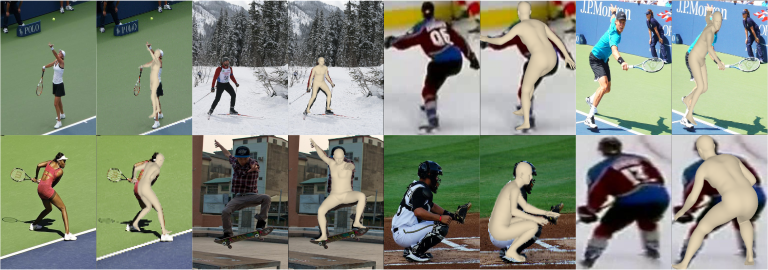}
  \vspace{-18px}
  \caption{\textbf{Qualitative results.} Qualitative comparison of D2A-HMR 2.0 on COCO and sports datasets with unusual poses.}
  \label{fig:d2ahmr2.0}
\end{figure*}

\paragraph{D2A-HMR 2.0.} Figure \ref{fig:d2ahmr2.0} presents qualitative results obtained by D2A-HMR 2.0 on various outdoor activities. These visualizations demonstrate the model's capability to achieve accurate alignment with input images, even in complex real-world scenarios. This highlights the effectiveness and robustness of D2A-HMR 2.0 for handling diverse outdoor environments.




\section{Limitations} \label{app:limitations}

\begin{figure}[H]
  \centering
  \includegraphics[width=\linewidth]{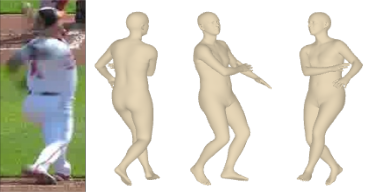}
  \vspace{-18px}
  \caption{\textbf{Limitations of the work.} The 3D human model falters to estimate the mesh vertices with severe motion blur.}
  \label{fig:limitation}
\end{figure}

PitcherNet, like many video analysis systems, is susceptible to error accumulation due to its reliance on a chain of interconnected components. Each step, from player identification to pitch analysis, introduces a degree of error. These errors can propagate throughout the processing pipeline, potentially leading to inaccuracies in the final extracted statistics. For instance, as shown in Figure \ref{fig:limitation}, motion blur during fast pitching can hinder the ability of the 3D human model (e.g., D2A-HMR 2.0) to accurately estimate joint positions, particularly in the pitching hand. This, in turn, significantly affects the performance of the extracted pitch statistics. 

To address the issue of severe motion blur and self-occlusion, we propose investigating strategies such as part-based regression (inspired by works such as \cite{pare}). This enhancement aims to better equip the model to effectively handle challenging conditions characterized by occlusion and motion blur.

\end{document}